\documentclass{article}
\usepackage{spconf,amsmath,amssymb,graphicx}
\usepackage[hidelinks]{hyperref}
\hypersetup{
pdftitle={P-SRM: Selective Recovery of Rejected Predictions in Visual Tracking},
pdfauthor={Youbin He; Siwei Wang},
pdfsubject={Visual tracking and post-rejection selective recovery},
pdfkeywords={Visual tracking, selective recovery, candidate verification, rejection mechanisms}}

\usepackage{booktabs,multirow}
\usepackage{etoolbox}
\patchcmd{\section}{-3.5ex}{-2.5ex}{}{}
\patchcmd{\section}{2.3ex}{1.8ex}{}{}
\patchcmd{\subsection}{-3.25ex}{-2.5ex}{}{}
\makeatletter
\long\def\@makecaption#1#2{%
 \vskip\abovecaptionskip
 \setbox\@tempboxa\hbox{#1. #2}%
 \ifdim\wd\@tempboxa>\hsize #1. #2\par
 \else\hbox to\hsize{\hfil\box\@tempboxa\hfil}\fi}
\makeatother

\newcommand{\concat}{\mathbin{\Vert}}

\DeclareMathOperator{\vecop}{vec}

\begin{document}
\raggedbottom
\title{P-SRM: Selective Recovery of Rejected Predictions in Visual Tracking}
\twoauthors{Youbin He}
{The Hong Kong Polytechnic University\\25091865d@connect.polyu.hk}
{Siwei Wang}
{The University of Hong Kong\\u3656944@connect.hku.hk}
\maketitle
\begin{abstract}
Many visual tracking methods use rejection mechanisms to suppress unreliable predictions. However, these mechanisms can also reject correctly localized candidates, leaving useful information unused. We investigate how to identify and recover these candidates while preserving native accepted outputs and candidate coordinates. To this end, we propose P-SRM (Post-rejection Selective Recovery Method), which combines spatial responses, past accepted states, and native decision margins to reassess candidates and selectively restore reliable predictions. We evaluate P-SRM on six trackers and four datasets spanning category-specific, point, and generic object tracking. Across all nine configurations, P-SRM improves rejected-candidate ranking and overall tracking performance. These results show that post-rejection verification can identify and recover useful predictions discarded by native rejection, demonstrating the value of reusing rejected information.
Project repository: \url{https://github.com/PalestyHR/P-SRM}.
\end{abstract}

\begin{figure*}[t]
\centering
\includegraphics[width=\textwidth]{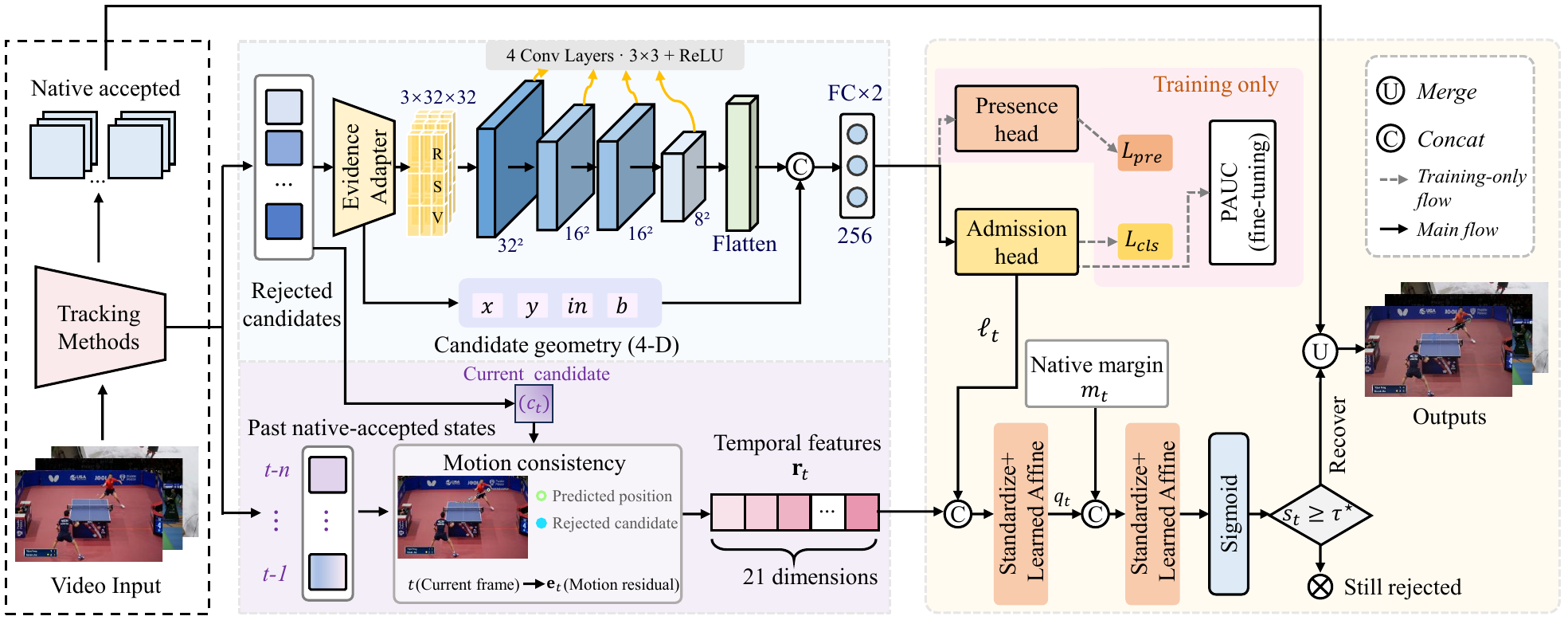}
\caption{Overview of P-SRM. Rejected candidates are selectively recovered using spatial, historical, and native decision evidence. Dashed arrows indicate training-only paths.}
\label{fig:psrm_overview}
\end{figure*}

\section{Introduction}
\label{sec:introduction}

Visual tracking estimates target locations across video frames. Many tracking methods use explicit rejection mechanisms to filter unreliable predictions. The TrackNet family targets small, fast-moving sports objects, including tennis balls and shuttlecocks. It decodes positions from heatmaps and suppresses weak responses before reporting locations~\cite{chen2023tracknet,huang2019tracknet,sun2020tracknetv2}. TAP-Net and TAPIR track specified points on physical surfaces and use visibility or localization uncertainty to judge prediction validity~\cite{doersch2022tapvid,doersch2023tapir}. KCF locates generic objects using correlation-filter responses~\cite{henriques2015kcf}; the implementation used here accepts or rejects predictions based on the response peak. These mechanisms serve their respective tasks by separating accepted outputs from rejected candidates.

Native rejection, however, does not necessarily imply incorrect localization. For example, a tracker may correctly locate a fast-moving shuttlecock in a blurred frame, yet withhold that position because its confidence score or heatmap peak falls below the threshold. Recovering this correctly localized but rejected candidate can restore a missing output. Yet rejected candidates also include mislocalizations and predictions made when the target is invisible. The key challenge is therefore to distinguish correctly localized candidates from these actual errors.

Related work addresses the distinction between reliable predictions and errors. Confidence calibration aligns predicted probabilities with correctness likelihood~\cite{guo2017calibration}. Mask Scoring R-CNN and MFTIQ estimate the quality of masks and flow correspondences, respectively~\cite{huang2019mask,serych2025mftiq}. ByteTrack uses low-score detections through association~\cite{zhang2022bytetrack}, while SelectiveNet jointly learns prediction and rejection~\cite{geifman2019selectivenet}. These approaches aim to improve confidence reliability, quality assessment, or prediction association and selection. We instead study candidate verification after native rejection: can the original tracker's evidence identify and recover correctly localized candidates while preserving native accepted outputs and candidate coordinates?

To answer this question, we propose P-SRM (Post-rejection Selective Recovery Method). P-SRM combines adapted spatial evidence, motion history, and native decision margins to reassess rejected candidates and selectively readmit reliable predictions (Fig.~\ref{fig:psrm_overview}). Our experiments confirm that native rejection does discard correctly localized candidates that can be identified and recovered. The results further show that recovering these candidates can also improve tracking performance.

Overall, our contributions include: (i) we identify correctly localized predictions excluded by native rejection, distinguishing admission errors from localization failures; (ii) we develop P-SRM to recover these candidates while preserving native accepted outputs and candidate coordinates; and (iii) we validate its identification and tracking benefits across six trackers and four datasets.

\begin{table*}[t]
\centering
\caption{Tracking results (\%). RTX 4090 inference (KCF: CPU). $N_C$: correct rejected candidates; $\bar n_C$: three-seed mean correct recoveries (recovery \%). $\ddagger$: TAP-Net FPS uses clip-amortized latency. Overhead includes clip-to-frame evidence preparation; separate profiling measures 3.94 ms/frame for preparation and 1.08 for core recovery.}
\label{tab:main_results}
\fontsize{9}{10.5}\selectfont
\setlength{\tabcolsep}{1pt}
\begin{tabular*}{\textwidth}{@{\extracolsep{\fill}}lllrrrrrrrrrrr@{}}
\toprule
Method / Source & Dataset & Setting & Acc. & Pre. & Rec. & F1 & AP$_r$ & AJ & OA & $N_C$ & \shortstack{$\bar n_C$\\(recovery \%)} & FPS & \shortstack{Extra latency\\(ms/frame)} \\
\midrule
\multicolumn{14}{l}{\textit{Category-specific tracking}}\\
\multirow{4}{*}{\shortstack[l]{TrackNetV1\\AVSS 2019~\cite{huang2019tracknet}}} & \multirow{2}{*}{Shuttlecock} & Native & 52.06 & \textbf{88.99} & 46.48 & 61.06 & 62.83 & -- & -- & \multirow{2}{*}{125} & & 25.3 & \multirow{4}{*}{7.14} \\
 &  & +P-SRM & \textbf{52.89} & 88.14 & \textbf{47.92} & \textbf{62.09} & \textbf{73.72} & -- & -- &  & 124.3 (99.47) & 21.5 &  \\
\cmidrule(l){2-13}
 & \multirow{2}{*}{RacketVision} & Native & 55.40 & \textbf{79.57} & 59.18 & 67.88 & 49.21 & -- & -- & \multirow{2}{*}{23} & & 25.3 &  \\
 &  & +P-SRM & \textbf{55.61} & 78.79 & \textbf{60.06} & \textbf{68.16} & \textbf{54.02} & -- & -- &  & 22.0 (95.65) & 21.5 &  \\
\cmidrule(l){1-14}
\multirow{4}{*}{\shortstack[l]{TrackNetV2\\ICPAI 2020~\cite{sun2020tracknetv2}}} & \multirow{2}{*}{Shuttlecock} & Native & 83.62 & \textbf{97.60} & 82.35 & 89.33 & 53.92 & -- & -- & \multirow{2}{*}{771} & & 39.8 & \multirow{4}{*}{4.27} \\
 &  & +P-SRM & \textbf{84.42} & 97.30 & \textbf{83.57} & \textbf{89.91} & \textbf{66.08} & -- & -- &  & 113.3 (14.70) & 34.0 &  \\
\cmidrule(l){2-13}
 & \multirow{2}{*}{RacketVision} & Native & 64.89 & \textbf{87.45} & 67.62 & 76.27 & 35.70 & -- & -- & \multirow{2}{*}{365} & & 39.8 &  \\
 &  & +P-SRM & \textbf{65.26} & 87.23 & \textbf{68.28} & \textbf{76.60} & \textbf{51.38} & -- & -- &  & 25.3 (6.94) & 34.0 &  \\
\cmidrule(l){1-14}
\multirow{4}{*}{\shortstack[l]{TrackNetV3\\MM Asia 2023~\cite{chen2023tracknet}}} & \multirow{2}{*}{Shuttlecock} & Native & 92.25 & \textbf{98.24} & 92.45 & 95.25 & 54.40 & -- & -- & \multirow{2}{*}{392} & & 8.5 & \multirow{4}{*}{5.30} \\
 &  & +P-SRM & \textbf{92.37} & 98.11 & \textbf{92.71} & \textbf{95.33} & \textbf{55.48} & -- & -- &  & 20.7 (5.27) & 8.2 &  \\
\cmidrule(l){2-13}
 & \multirow{2}{*}{RacketVision} & Native & 75.03 & \textbf{88.59} & 80.47 & 84.33 & 60.48 & -- & -- & \multirow{2}{*}{394} & & 8.5 &  \\
 &  & +P-SRM & \textbf{76.75} & 88.58 & \textbf{82.81} & \textbf{85.60} & \textbf{74.00} & -- & -- &  & 106.7 (27.07) & 8.2 &  \\
\midrule
\multicolumn{14}{l}{\textit{Point tracking}}\\
\multirow{2}{*}{\shortstack[l]{TAP-Net\\NeurIPS 2022~\cite{doersch2022tapvid}}} & \multirow{2}{*}{TAP-Vid Kinetics} & Native & 80.37 & 75.36 & 91.03 & 82.46 & 19.92 & 38.68 & 79.62 & \multirow{2}{*}{54,759} & & 517.3$\ddagger$ & \multirow{2}{*}{4.99$\ddagger$} \\
 &  & +P-SRM & \textbf{80.81} & \textbf{75.38} & \textbf{92.29} & \textbf{82.98} & \textbf{52.05} & \textbf{39.13} & \textbf{80.29} &  & 7,657.3 (13.98) & 144.4$\ddagger$ &  \\
\cmidrule(l){1-14}
\multirow{2}{*}{\shortstack[l]{Online TAPIR\\ICCV 2023~\cite{doersch2023tapir}}} & \multirow{2}{*}{TAP-Vid Kinetics} & Native & 89.83 & \textbf{88.57} & 93.23 & 90.84 & 34.91 & 50.12 & 83.93 & \multirow{2}{*}{44,115} & & 54.9 & \multirow{2}{*}{1.90} \\
 &  & +P-SRM & \textbf{89.92} & 88.16 & \textbf{94.00} & \textbf{90.99} & \textbf{42.55} & \textbf{50.29} & \textbf{84.35} &  & 5,022.7 (11.39) & 49.7 &  \\
\midrule
\multicolumn{14}{l}{\textit{Generic object tracking}}\\
\multirow{2}{*}{\shortstack[l]{KCF\\TPAMI 2015~\cite{henriques2015kcf}}} & \multirow{2}{*}{OTB2013} & Native & 32.72 & \textbf{87.31} & 34.34 & 49.29 & 43.50 & -- & -- & \multirow{2}{*}{2,552} & & 252.7 & \multirow{2}{*}{0.82} \\
 &  & +P-SRM & \textbf{33.51} & 86.82 & \textbf{35.29} & \textbf{50.18} & \textbf{45.04} & -- & -- &  & 279.7 (10.96) & 209.4 &  \\
\bottomrule
\end{tabular*}
\end{table*}

\section{Method}
\label{sec:method}

P-SRM applies to trackers that expose an explicit validity decision together with rejected candidate locations and spatial evidence (Fig.~\ref{fig:psrm_overview}).

\subsection{Problem setting and recoverable candidates}
\label{sec:candidates}

For candidate $\mathbf c_t$ at frame $t$, $d_t=1$ denotes native acceptance and $d_t=0$ rejection. Within the rejected set, we distinguish three classes: $C$, visible and correctly localized; $L$, visible but mislocalized; and $A$, the target is annotated as invisible.
Point candidates are correct when $\|\mathbf c_t-\mathbf c_t^\star\|_2\leq\epsilon$, where $\mathbf c_t^\star$ is the annotation and $\epsilon$ the localization tolerance. Box candidates require an intersection-over-union of at least $0.5$ with the annotated box.

With candidate coordinates fixed, $C$ is the recoverable class. Annotations define the correctness target $y_t=1$ for $C$ and $y_t=0$ otherwise; inference estimates correctness from available evidence.

\subsection{Spatial evidence adaptation and quality learning}
\label{sec:adapter}
\label{sec:learning}

Identifying recoverable candidates requires assessing target presence and its spatial agreement with the candidate. The adapter aligns $\mathbf c_t$ with evidence $E_t$ from the same native forward pass, producing response plane $R_t$, candidate support $S_t$, and valid support $V_t$. These encode the response distribution, candidate location, and usable image regions, respectively:
\begin{equation}
(X_t,\mathbf g_t)=\mathcal A_b(E_t,\mathbf c_t),\quad
X_t=R_t\concat S_t\concat V_t.
\label{eq:adapter}
\end{equation}
Here $b$ identifies the tracker and $\concat$ denotes concatenation. Geometry $\mathbf g_t$ contains normalized candidate coordinates and in-frame and support-boundary indicators.

Encoder $f_\theta$ extracts response patterns; projection $g_\theta$ combines them with candidate geometry for the admission head~\cite{huang2019mask}:
\begin{equation}
\mathbf z_t=g_\theta\!\left(\vecop(f_\theta(X_t))\concat\mathbf g_t\right),\qquad
\ell_t=\mathbf w_q^\top\mathbf z_t+b_q.
\label{eq:quality}
\end{equation}
The quality score $p_t=\sigma(\ell_t)$ is trained against $y_t$.

A training-only presence head $u_t=\mathbf w_a^\top\mathbf z_t+b_a$ complements correctness supervision with visibility labels $a_t=1$ for $C\cup L$ and $a_t=0$ for $A$:
\begin{equation}
\mathcal L_{\mathrm{init}}=
\mathbb E_{t\in\mathcal R_{\mathrm{tr}}}
\left[\mathcal L_{\mathrm{cls},t}+\lambda_a\mathcal L_{\mathrm{pre},t}\right].
\label{eq:initial_loss}
\end{equation}
Here $\mathcal L_{\mathrm{cls},t}=B_{w_y}(p_t,y_t)$ and $\mathcal L_{\mathrm{pre},t}=B_{w_a}(\sigma(u_t),a_t)$ are admission and presence losses. $B_w$ is positive-class-weighted binary cross-entropy on rejected training set $\mathcal R_{\mathrm{tr}}$, with auxiliary weight $\lambda_a$. Only the admission logit enters fusion.

Ranking refinement uses positive and negative sets $\mathcal P,\mathcal N$ defined by $y_t$. Pairwise loss $L_{ij}=[\kappa-(p_i-p_j)]_+^2$, for $i\in\mathcal P,j\in\mathcal N$, encourages higher scores for correct candidates. The KL-DRO partial-AUC (pAUC) objective~\cite{zhu2022auc} emphasizes larger pairwise errors:
\begin{equation}
\mathcal L_{\mathrm{pAUC}}=
\frac{\Lambda}{|\mathcal P|}\sum_{i\in\mathcal P}
\log\!\left[\frac{1}{|\mathcal N|}\sum_{j\in\mathcal N}
\exp\!\left(\frac{L_{ij}}{\Lambda}\right)\right].
\label{eq:tail_loss}
\end{equation}
Here $[x]_+=\max(x,0)$, $\kappa$ is the desired score separation, and $\Lambda$ controls the emphasis on larger pairwise errors.

\subsection{History and native decision fusion}
\label{sec:temporal}

To complement spatial evidence with motion consistency, P-SRM uses past accepted states $\mathcal H_t=\{(k,\mathbf c_k,\boldsymbol\xi_k):k<t,d_k=1\}$, where $\boldsymbol\xi_k$ stores confidence information.

When at least two past acceptances are available, the latest positions at $t_1<t_2<t$ provide a velocity estimate. Extrapolating it to the current frame gives the candidate's deviation from the expected position:
\begin{equation}
\mathbf v_t=\frac{\mathbf c_{t_2}-\mathbf c_{t_1}}{t_2-t_1},\qquad
\mathbf e_t=\mathbf c_t-\mathbf c_{t_2}-(t-t_2)\mathbf v_t.
\label{eq:innovation}
\end{equation}
History descriptor $\mathbf r_t=\psi(\mathbf c_t,\mathcal H_t)$ combines this deviation, time gaps, and confidence statistics. Unavailable terms are zero-filled before standardization, with flags for zero, one, or at least two past acceptances.

Native margin $m_t$ indicates the candidate's position relative to the native admission boundary. Fusion combines spatial quality logit $\ell_t$ with history $\mathbf r_t$, then incorporates $m_t$:
\begin{equation}
\begin{aligned}
q_t&=\mathbf a^\top T_1(\ell_t\concat\mathbf r_t)+b_1,\\
s_t&=\sigma\!\left(\mathbf b^\top T_2(q_t\concat m_t)+b_2\right).
\end{aligned}
\label{eq:fusion}
\end{equation}

With the quality network fixed, the readouts are fitted sequentially using video-grouped five-fold cross-fitting and $L_2$-regularized, class-balanced binary cross-entropy against $y_t$. The second uses the first's out-of-fold logits. Standardization transforms $T_1,T_2$ estimate means and standard deviations from each fitting partition. Readouts and transforms are refitted on all source fitting data and fixed for external evaluation.

\subsection{Selective recovery}
\label{sec:recovery}

We select $\tau^\star$ from pooled out-of-fold scores of the second readout to maximize correct recoveries under the predefined source-domain calibration rules. The threshold is then fixed, preserving native acceptances and readmitting rejected candidates when $s_t\geq\tau^\star$:
\begin{equation}
\widehat d_t=\begin{cases}
1, & d_t=1,\\[2pt]
\mathbb I[s_t\geq\tau^\star], & d_t=0,
\end{cases}
\label{eq:admission}
\end{equation}
where $\mathbb I[\cdot]$ is the indicator function. Admitted candidates retain their original coordinates.

\begin{figure*}[t]
\centering
{\fontsize{9}{10.5}\selectfont\noindent
\hspace*{.025\textwidth}%
\makebox[.11425\textwidth]{Native}\makebox[.11425\textwidth]{+P-SRM}\hspace{.012\textwidth}%
\makebox[.11425\textwidth]{Native}\makebox[.11425\textwidth]{+P-SRM}\hspace{.012\textwidth}%
\makebox[.11425\textwidth]{Native}\makebox[.11425\textwidth]{+P-SRM}\hspace{.012\textwidth}%
\makebox[.11425\textwidth]{Native}\makebox[.11425\textwidth]{+P-SRM}\par}
\includegraphics[width=\textwidth]{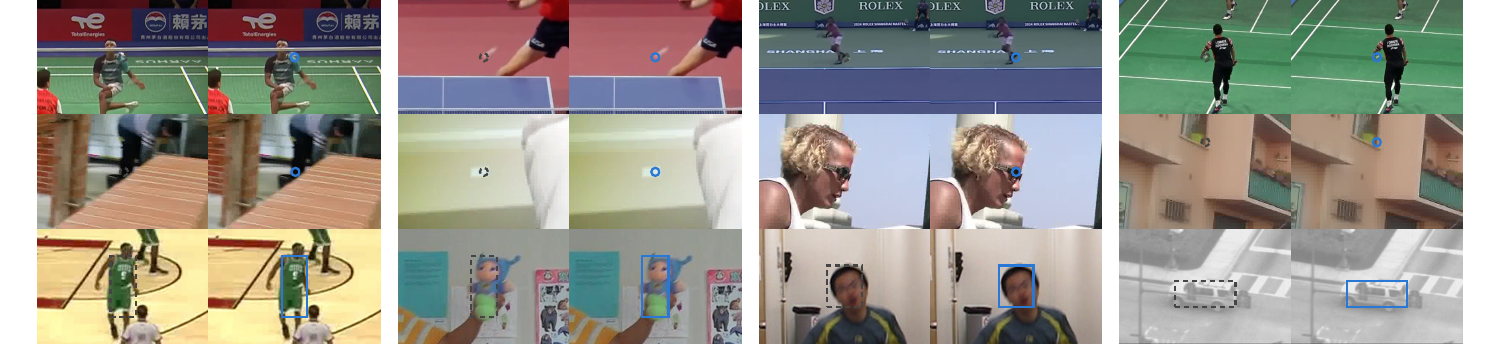}
\caption{Recovery examples across three tracking categories. Gray dashed markers show native-rejected candidates; blue markers show the same candidates after recovery. Examples are from RacketVision, Shuttlecock, Kinetics, and OTB2013.}
\label{fig:recovery_examples}
\end{figure*}

\section{Experiments and Results}
\label{sec:experiments}

\subsection{Benchmarks and evaluation protocol}

We evaluate P-SRM in category-specific, point, and generic object tracking, covering six trackers and four datasets (Table~\ref{tab:main_results})~\cite{dong2026racket,doersch2022tapvid,wu2013otb}. TrackNet and point-tracking configurations use source data for fitting and calibration, followed by evaluation on separate splits. KCF uses video-grouped five-fold evaluation on OTB2013 with supplementary visibility annotations.

Native predictions are fixed. P-SRM is fitted and OOF-calibrated for each tracker configuration. Main results average seeds 42, 3407, and 8008; cue ablations in Table~\ref{tab:paired_recovery} use seed 42.

AP$_r$ is the primary metric for identifying correct candidates within the native rejection set. Final tracking metrics measure the benefits realized after selective admission. Accuracy, precision, recall, and F1 pool events; Average Jaccard (AJ) and Occlusion Accuracy (OA) average videos.

\subsection{Rejected-candidate identification and recovery}

Table~\ref{tab:main_results} shows that every native rejection set contains correctly localized candidates. P-SRM identifies and recovers these candidates without changing their coordinates, improving AP$_r$ across all nine configurations. The 95\% paired-bootstrap confidence intervals for pooled F1 gains over Native also lie entirely above zero in every configuration.

Even in configurations with low correct-candidate recovery rates, AP$_r$ improves substantially. This finding highlights the value of assessing rejected-candidate quality separately from the final admission decision.

\subsection{Ablation and sensitivity analysis}

To assess the need for full P-SRM, we compare simpler recovery schemes using quality (Q), history (H), and native margin (M): M, Q, M+H, Q+H, and Q+M (Table~\ref{tab:paired_recovery}). M uses only the native scalar margin through a one-dimensional logistic readout calibrated on source-domain OOF scores. For both trackers, the final fitted mapping is increasing, making its admission rule equivalent to native-score thresholding while preserving native accepted outputs. All schemes share candidates, source partitions, and calibration rules, with separately calibrated thresholds.

\begin{table}[t]
\centering
\caption{Cue ablations (pp); CIs: 10,000 paired video bootstraps with fixed models/thresholds. $\Delta t$: five-run mean end-to-end change (ms/frame, RTX 4090).}
\label{tab:paired_recovery}
\fontsize{9}{10.5}\selectfont
\setlength{\tabcolsep}{1.2pt}
\begin{tabular*}{\columnwidth}{@{\extracolsep{\fill}}lrrrr@{}}
\toprule
Comparison & $\Delta t$ & $\Delta$AP$_r$ & $\Delta$F1$_{\rm pool}$ & $\Delta$F1$_{\rm video}$ [95\% CI]\\
\midrule
\multicolumn{5}{l}{\textit{TAP-Net}}\\
$Q\!\to\!Q{+}H$ & 1.00 & 3.92 & 0.020 & $0.017\;[-0.046, 0.076]$ \\
$Q{+}H\!\to\!\mathrm{Full}$ & 0.30 & 0.11 & -0.011 & $-0.016\;[-0.024, 0.007]$ \\
$Q{+}M\!\to\!\mathrm{Full}$ & 1.36 & 4.04 & -0.001 & $-0.010\;[-0.070, 0.044]$ \\
$M\!\to\!\mathrm{Full}$ & 6.26 & 32.89 & 0.588 & $0.917\;[0.620, 1.258]$ \\
$M{+}H\!\to\!\mathrm{Full}$ & 5.51 & 20.16 & 0.508 & $0.715\;[0.460, 1.008]$ \\
\midrule
\multicolumn{5}{l}{\textit{Online TAPIR}}\\
$Q\!\to\!Q{+}H$ & 1.08 & 4.66 & 0.026 & $0.028\;[-0.014, 0.065]$ \\
$Q{+}H\!\to\!\mathrm{Full}$ & 0.53 & 1.48 & 0.042 & $0.050\;[0.001, 0.092]$ \\
$Q{+}M\!\to\!\mathrm{Full}$ & 0.98 & 2.34 & 0.048 & $0.050\;[0.012, 0.083]$ \\
$M\!\to\!\mathrm{Full}$ & 2.84 & 7.89 & 0.146 & $0.200\;[0.127, 0.273]$ \\
$M{+}H\!\to\!\mathrm{Full}$ & 1.08 & 5.25 & 0.092 & $0.122\;[0.063, 0.185]$ \\
\bottomrule
\end{tabular*}
\end{table}

\begin{table}[t]
\centering
\caption{Hyperparameter sensitivity analysis. Defaults: $(\lambda_a,\kappa,\Lambda)=(0.5,0.6,1)$. Seed: 3407.}
\label{tab:hyperparam}
\fontsize{9}{10.5}\selectfont
\setlength{\tabcolsep}{2pt}
\begin{tabular*}{\columnwidth}{@{\extracolsep{\fill}}lrrrr@{}}
\toprule
 & \multicolumn{2}{c}{TAP-Net} & \multicolumn{2}{c}{TrackNetV3}\\
Setting & AP$_r$ (\%) & F1 (\%) & AP$_r$ (\%) & F1 (\%)\\
\midrule
Default & 51.27 & 82.85 & 55.22 & 95.34 \\
\midrule
$\lambda_a=0.25$ & 51.18 & 83.17 & 55.20 & 95.35 \\
$\lambda_a=1.0$ & 47.29 & 82.91 & 55.18 & 95.36 \\
$\kappa=0.4$ & 51.18 & 82.57 & 55.20 & 95.35 \\
$\kappa=0.8$ & 51.33 & 82.89 & 55.23 & 95.34 \\
$\Lambda=0.5$ & 51.27 & 82.68 & 55.21 & 95.34 \\
$\Lambda=2.0$ & 51.24 & 82.90 & 55.23 & 95.34 \\
\bottomrule
\end{tabular*}
\end{table}

Full outperforms M and M+H in ranking and recovery for both point trackers, demonstrating the value of spatial quality information not fully exploited by native scores and history. Comparisons with Q, Q+H, and Q+M further support the cues' complementary value for ranking. It should be clarified that the complementary value of these cues for ranking does not imply higher final tracking metrics in every setting. Because trackers process candidates differently, the contribution of the same cue may vary across settings, allowing some simpler combinations to achieve better results in certain cases. We nevertheless adopt Full for its applicability across tracking settings and demonstrated identification and tracking gains.

We additionally assess sensitivity to training hyperparameters by varying one parameter at a time in TAP-Net/Kinetics and TrackNetV3/Shuttlecock. Tracking performance remains broadly stable across the tested settings, with consistent gains over Native (Table~\ref{tab:hyperparam}).

\section{Conclusion}
\label{sec:conclusion}

P-SRM demonstrates that correctly localized predictions can be recovered after native rejection. Across multiple trackers and datasets, this improves candidate identification and tracking performance, establishing post-rejection verification as a practical way to reuse existing predictions.

\clearpage
\section*{\centering\normalsize COMPLIANCE WITH ETHICAL STANDARDS}
All experiments use existing, publicly available datasets: Shuttlecock, RacketVision, TAP-Vid Kinetics, and OTB2013. We use dataset-provided annotations, including the corrected Shuttlecock Test labels released with TrackNetV3. For OTB2013, we retain the original bounding-box annotations and add supplementary target-visibility labels. This study involves no new participant recruitment or video recording.

\section*{\centering\normalsize ACKNOWLEDGMENTS}
This work was funded personally by the first author. No external funding was received. The authors declare no competing interests.

\patchcmd{\thebibliography}{\section{References}}{\section*{\centering\normalsize REFERENCES}}{}{}
\bibliographystyle{IEEEbib}
\bibliography{references}
\end{document}